\documentclass[preprint,12pt]{elsarticle}

\usepackage{amssymb}

\usepackage{amsmath}
\usepackage{graphicx}
\usepackage{xcolor}
\usepackage{soul}
\usepackage[colorlinks=true, allcolors=blue]{hyperref}

\usepackage[draft]{changes}
\definechangesauthor[color=red]{gm}
\definechangesauthor[color=blue]{yunfei}
\definechangesauthor[color=purple]{yingya}
\definechangesauthor[color=yellow]{gp}
\usepackage{multirow}
\usepackage{graphicx}
\usepackage{booktabs}
\usepackage{float}
\usepackage{makecell}
\usepackage{longtable}
\usepackage{lscape}
\journal{review}

\begin{document}

\begin{frontmatter}



\title{From Explainable to Interpretable Deep Learning for Natural Language Processing in Healthcare: How Far from Reality?}


\author[label1]{Guangming Huang}
\ead{gh22231@essex.ac.uk}
\author[label2]{Yingya Li}
\ead{yingya.li@childrens.harvard.edu}
\author[label3]{Shoaib Jameel}
\ead{m.s.jameel@southampton.ac.uk}
\author[label1]{Yunfei Long\corref{cor1}}
\ead{yl20051@essex.ac.uk}
\author[label4]{Giorgos Papanastasiou\corref{cor1}}
\ead{g.papanastasiou@athenarc.gr}
\cortext[cor1]{Corresponding author}

\affiliation[label1]{organization={School of Computer Science and Electronic Engineering, University of Essex},
            city={Colchester},
            postcode={CO4 3SQ}, 
            country={United Kingdom}}
\affiliation[label2]{organization={Harvard Medical School and Boston Children’s Hospital},
            city={Boston},
            postcode={02115}, 
            country={United States}}
\affiliation[label3]{organization={Electronics and Computer Science, University of Southampton},
            city={Southampton},
            postcode={SO17 1BJ}, 
            country={United Kingdom}}
\affiliation[label4]{organization={Archimedes Unit, Athena Research Centre},
            city={Athens},
            postcode={15125}, 
            country={Greece}}

\begin{abstract}
Deep learning (DL) has substantially enhanced natural language processing (NLP) in healthcare research. However, the increasing complexity of DL-based NLP necessitates transparent model interpretability, or at least explainability, for reliable decision-making. This work presents a thorough scoping review of explainable and interpretable DL in healthcare NLP. The term ``eXplainable and Interpretable Artificial Intelligence'' (XIAI) is introduced to distinguish XAI from IAI. Different models are further categorized based on their functionality (model-, input-, output-based) and scope (local, global). Our analysis shows that attention mechanisms are the most prevalent emerging IAI technique. The use of IAI is growing, distinguishing it from XAI. The major challenges identified are that most XIAI does not explore ``global'' modelling processes, the lack of best practices, and the lack of systematic evaluation and benchmarks. One important opportunity is to use attention mechanisms to enhance multi-modal XIAI for personalized medicine. Additionally, combining DL with causal logic holds promise. Our discussion encourages the integration of XIAI in Large Language Models (LLMs) and domain-specific smaller models. In conclusion, XIAI adoption in healthcare requires dedicated in-house expertise. Collaboration with domain experts, end-users, and policymakers can lead to ready-to-use XIAI methods across NLP and medical tasks. While challenges exist, XIAI techniques offer a valuable foundation for interpretable NLP algorithms in healthcare.
\end{abstract}




\end{frontmatter}


\section{Introduction}
Recently, deep learning (DL) has been instrumental in enhancing the performance of natural language processing (NLP) in a wide range of healthcare settings \cite{nguyen2022improving, koleck2019natural, kim2022can, zhang2022section, cahyawijaya2022long, michalopoulos2022medicalsum, wu2022deltanet, you2022jpg, moro2022discriminative, otmakhova2022patient}. DL has notably advanced the analysis of NLP in areas like medical diagnosis \cite{grundmann2022attention, yan2022clinical, wu2022deltanet}, patient monitoring \cite{liu2019fast} and drug discovery \cite{yang2022ddi, iinuma2022improving}.

The evolution of DL models in NLP has led to substantial performance improvements over traditional machine learning methods, with advancements ranging from recurrent neural networks (RNNs) \cite{rumelhart1986learning} and long short-term memory (LSTM) networks \cite{hochreiter1997long} to convolutional neural networks (CNNs) \cite{kim2014convolutional} and the latest self-attention mechanisms and transformers \cite{vaswani2017attention}. However, these gains in performance come at the cost of increasingly complex model architectures. In healthcare settings, where trustworthy and transparent decision-making is paramount, there is a growing need for interpretable or explainable models \cite{danilevsky2020survey}. Both explainability and interpretability pose significant challenges, as understanding how NLP embeddings translate into DL decision-making mechanisms remains complex \cite{sun2021interpreting, payrovnaziri2020explainable}.  Fortunately, recent research in explainable \cite{ozyegen2022word, thorsen2022discrete} and interpretable \cite{teng2020explainable, dong2021explainable} DL for NLP in healthcare shows promise. Furthermore, the rise of large language models (LLMs) highlights the growing importance of evaluating which explainable and interpretable methods are most beneficial for healthcare, especially as data and model complexity increase over time.

In our work, the evolution of eXplainable and Interpretable Artificial Intelligence (denoted with the new term ``XIAI'') is reviewed \cite{lipton2018mythos, doshi2017towards, carvalho2019machine, rudin2019stop}. Since there is no systematic definition of IAI and XAI \cite{lipton2018mythos, doshi2017towards}, there is an ongoing inconsistency regarding how these terms are used in DL \cite{carvalho2019machine}. To define XIAI in this review, the statistical definition by \citet{rudin2019stop} is followed: IAI focuses on designing inherently interpretable models; XAI aims to provide post hoc model explanations. Further, XIAI method designs were evaluated, by organizing them into model-based, input-based, and output-based approaches. To improve clarity and understanding, XIAI is further grouped based on their scope (local, global): local XIAI yields insights derived from particular inputs, whereas global XIAI grants a wider understanding based on the entire predictive mechanism of the model \cite{danilevsky2020survey}. Our analysis seeks to enhance clarity and insight into the most effective ways to combine DL and XIAI methods in healthcare NLP by examining and revealing essential architectural designs. Through a quantitative assessment of IAI and XAI in both NLP and medical tasks, we pinpoint current and future opportunities, challenges, advantages, and limitations. In line with the most recent LLM developments, our review evaluates whether attention mechanisms can reinforce the implementation of IAI against XAI. By carefully investigating XIAI DL in healthcare NLP, our review aims to provide insights into the scientific and clinical impact that can potentially be important for XIAI democratization in healthcare NLP.

This review is organised as follows. Initially, the key related work in the domain is reviewed by analysing the important XIAI research trends, methodology designs, paradigms and scope. Subsequently, our review analyses challenges and opportunities for translating XIAI into clinical practice. Finally, we assess the gap between current XIAI capabilities and real-world healthcare implementation, highlighting areas for future research.

\section{Methods}

\begin{figure}[]
\centering
\includegraphics[width=1\textwidth]{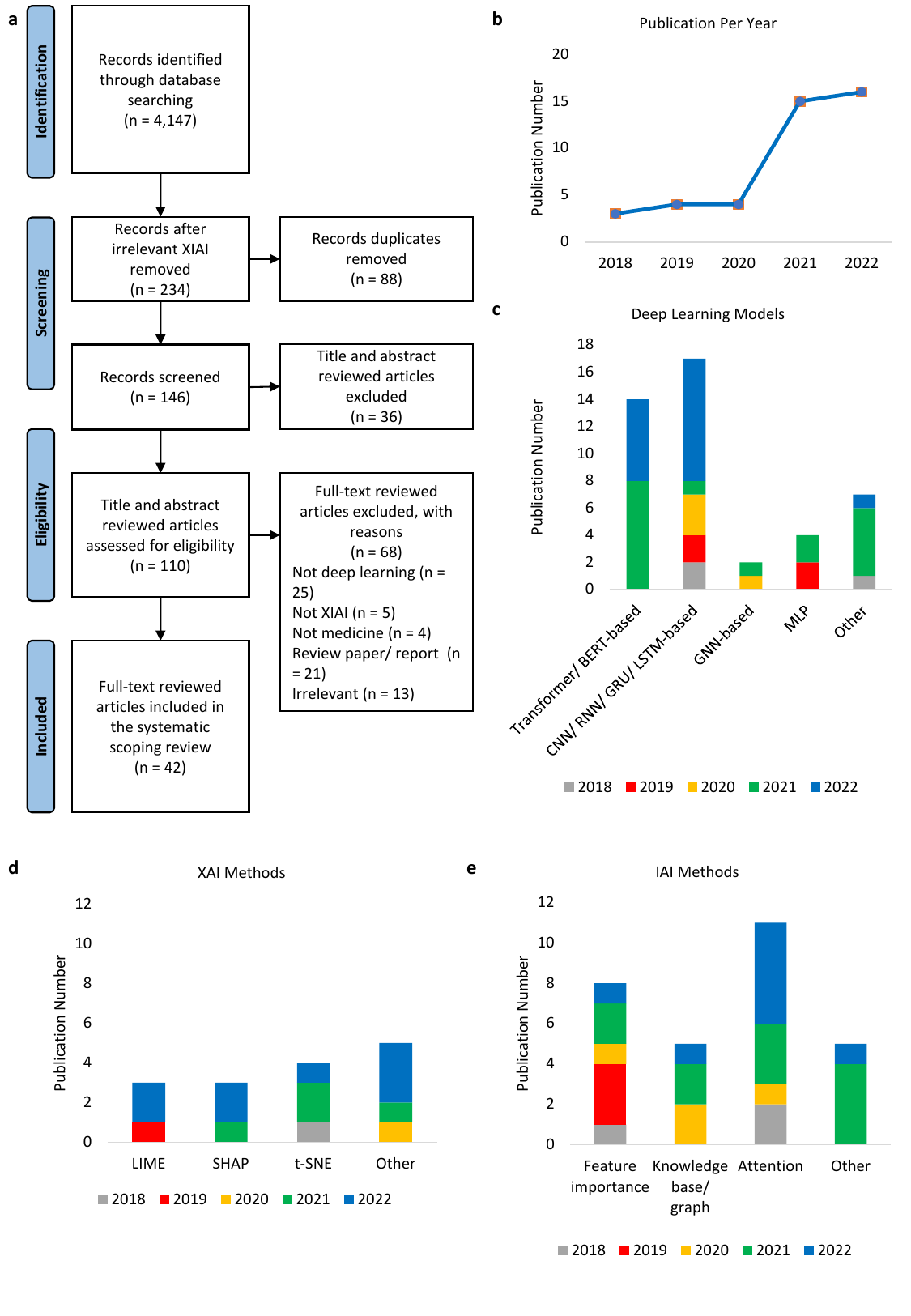}\caption{\label{fig:1} (a) PRISMA flowchart. The flowchart presents the inclusion and exclusion of papers at each review stage. (b) Publication records of XIAI, (c) main DL model category, (d) main XAI and (e) main IAI methods, per year.}
\end{figure}

\subsection{Literature review strategy}  
A comprehensive scoping review of explainable and interpretable deep learning (DL)--based natural language processing (NLP) techniques applied to healthcare was conducted, considering publications between January 1, 2018, and December 31, 2022. We searched the following databases: Scopus, Web of Science, PubMed, and the Association for Computing Machinery (ACM), following the Preferred Reporting Items for Systematic Reviews and Meta-Analyses (PRISMA) guidelines \cite{moher2009preferred}. To address real-world data needs and assess the potential for the democratization of XIAI methods, our review focused on electronic health record (EHR) and electronic medical record (EMR) NLP data. The emergence of Transformers as an effective approach to learning long-range interactions via attention modules in 2017 \cite{vaswani2017attention} marked the beginning of a new era in NLP, with Transformer variants becoming the most prevalent and promising techniques \cite{lin2022survey}. To focus our review on these significant developments, our observation window was selected to start in 2018. Figure~\ref{fig:1}(a) illustrates the PRISMA flow.

\textbf{Initial Filtering:} To broaden the search, publications were initially retrieved using the following keywords in the title, abstract, and manuscript keywords: \textit{(``natural language processing'' OR ``NLP'') AND ( ``healthcare'' OR ``medic'' OR ``clinic'' OR ``EHR'' OR ``EMR'')}. This led to a preliminary list of 4,147 papers. Subsequently, to refine the focus on XIAI methods, irrelevant papers were removed, resulting in 234 papers. In the subsequent step, removing duplicate records across databases resulted in 146 unique papers selected for title and abstract screening.

\textbf{Title and Abstract Screening:} All authors independently reviewed the titles and abstracts of all 146 papers. Articles demonstrably irrelevant to the field of study were excluded, resulting in 110 papers eligible for full-text review.

\textbf{Full-Text Review:} During the full-text review, 68 papers were excluded for the following reasons:

\begin{itemize}
    \item 25 papers did not utilize DL techniques.
    \item 5 papers were not relevant to XIAI.
    \item 4 papers lacked a medical focus.
    \item 21 papers were review articles.
    \item 13 papers fell outside the scope of DL, NLP, or healthcare.
\end{itemize}

This resulted in a final selection of 42 journal and conference papers for our review analysis.

\subsection{Review aspects} 

\textbf{Full-Text Evaluation:} During the full-text review, we focused on evaluating the following aspects of each paper:

\begin{enumerate}
    \item Publication Date: This provided context for the development of the methods.
    \item DL Model: The specific DL model employed was identified (see Results for DL model categorization).
    \item IAI vs. XAI: The paper's focus on interpretability or explainability was distinguished.
    \item Scope of XIAI Method: We categorized the XIAI method as local or global based on its ability to inform about the model predictions regarding a specific input, or the entire modelling process (end-to-end), respectively \cite{danilevsky2020survey}.
    \item XIAI Method Type: The type of XIAI method used (model-based, input-based, or output-based) was identified.
    \item Medical Task: The specific medical task addressed by the XIAI approach was categorized.
    \item NLP Task: The type of NLP task involved (e.g., classification, information extraction) was identified.
    \item Data Source: We categorized the data source used (public or private data).
    \item Source Code Availability: The availability of the source code for the method was assessed.
    \item XIAI Evaluation: The presence or absence of an evaluation for the XAI method was noted.
\end{enumerate}

Furthermore, we analyzed the XIAI methods based on three key aspects to identify challenges, opportunities, advantages, and drawbacks:

XAI and IAI methods with DL Models, NLP Tasks, and Medical Tasks: This analysis explored how XAI and IAI methods were combined with a) the specific DL model used, b) the type of NLP task analyzed, and c) the medical task addressed.

\section{Results}
The results section is organised as follows: first the research trends are detailed, then the XAI against IAI methods, the scope of XIAI methods, the three XIAI method paradigms (i.e., model-, input- and output-based methods), the DL models explained/ interpreted and the NLP and medical tasks addressed. Further, the evaluation metrics of XIAI methods and open code are described.  Finally, the dataset sources (public and/ or private) are assessed. 

\subsection{Research trends}

Published work on XAI and IAI NLP in healthcare was reviewed. A rapid increase in the number of studies was observed from 2021 to 2022, compared to the first years of our observation window (Figure~\ref{fig:1}(b)). This trend coincides with the dominance of Tranformer DL models (Figure~\ref{fig:1}(b)). In addition, this trend reflects the rapid rise of attention mechanisms, which have rapidly overtaken other XIAI methods during this time (Figure~\ref{fig:1}(d-e)).

\subsection{IAI against XAI}

All methods across papers were examined to assess whether they focused on IAI or XAI. More studies involved IAI methods ($n = 29$) than XAI methods ($n = 15$), with this trend being evident in the period 2020-2022. Figure \ref{fig:1}(d) illustrates the publication record of XAI methods used per year.

Among XAI methods, t-SNE ($n = 4$) was the most widely used, followed by LIME ($n = 3$), SHAP ($n = 3$) and other methods ($n = 5$) such as maximum aggregator (MaXi) \cite{martina2020classification}, SKET eXplained (SKET X) \cite{marchesin2022empowering}, syntax tree-guided semantic explanation (STEP) \cite{farruque2021explainable}, evidence-based \cite{de2022explainable} and sentiment intensity score \cite{boukobza2022deep}.

As illustrated in Figure \ref{fig:1}(e), most IAI methods involved attention mechanisms ($n = 11$), followed by feature importance ($n = 8$), knowledge base/graph ($n = 5$) and other methods ($n =5$), including causal graphs \cite{chen2021causal}, logistic regression-based parametric predictions \cite{ferte2021automatic}, opinion aggregator \cite{lu2021revealing}, case-based reasoning \cite{amador2021case} and interactive classification \cite{ahne2022improving}.

\subsection{Scope of XIAI methods}
XIAI method scopes were further grouped into local and global  \cite{danilevsky2020survey}. The majority of papers employed local methods ($n=37$),with only $n=5$ studies involving global XIAI (Table \ref{tab:1}). 

\subsection{Three XIAI method paradigms}
Three distinct XIAI categories were introduced: model-, input- and output-based methods. These definitions were conceptualized based on whether an XIAI method relies on internal/external modules (model) to perform XIAI, measures how the input features affect model decisions or explains/interprets model behaviour through analyzing prediction outcomes, respectively. Table \ref{tab:1} presents a comprehensive overview of sixteen different XIAI approaches identified in our review and their distinct categories (model-, input-, and output-based methods). There were eighteen, thirteen and thirteen model-, input- and output-based methods identified, respectively. 

According to Table \ref{tab:1}, the most popular model-based XIAI was t-SNE ($n = 4$), followed by LIME ($n = 3$), SHAP ($n = 3$) and other methods that have been sparsely used. Among input-based XIAI, studies either used feature importance-based ($n = 8$) or knowledge graph-based ($n = 5$) methods. Regarding output-based methods, the attention mechanism was the most frequently used approach (Table \ref{tab:1}). While benefiting from their performance advancements, attention mechanisms may therefore be an important IAI approach. 

\begin{table}[htb]
\centering
\resizebox{\textwidth}{!}{%
\renewcommand\arraystretch{1.5}
\begin{tabular}{|c|l|c|c|p{3cm}<{\centering}|}
\hline
\textbf{XIAI   category} &
  \multicolumn{1}{c|}{\textbf{XIAI methods}} &
  \textbf{Explainable/Interpretable} &
  \textbf{Local/Global} &
  \multicolumn{1}{c|}{\textbf{Articles}} \\ \hline
\multirow{11}{*}{Model-based} &
  Causal graph &
  Interpretable &
  Global &
  \cite{chen2021causal} \\ \cline{2-5} 
 &
  \makecell[l]{Logistic regression-based \\ parametric predictor} &
  Interpretable &
  Global &
  \cite{ferte2021automatic} \\ \cline{2-5} 
 &
  Opinion aggregator &
  Interpretable &
  Global &
  \cite{lu2021revealing} \\ \cline{2-5} 
 &
  Case-based reasoning &
  Interpretable &
  Global &
  \cite{amador2021case} \\ \cline{2-5} 
 &
  Interactive classification &
  Interpretable &
  Global &
  \cite{ahne2022improving} \\ \cline{2-5} 
 &
  LIME &
  Explainable &
  Local &
  \cite{zhang2019explainable, ozyegen2022word,   uddin2022depression} \\ \cline{2-5} 
 &
  MAXi &
  Explainable &
  Local &
  \cite{martina2020classification} \\ \cline{2-5} 
 &
  SHAP &
  Explainable &
  Local &
  \cite{thorsen2022discrete, maji2021interpretable,   naseem2022hybrid} \\ \cline{2-5} 
 &
  SKET X &
  Explainable &
  Local &
  \cite{marchesin2022empowering} \\ \cline{2-5} 
 &
  STEP &
  Explainable &
  Local &
  \cite{farruque2021explainable} \\ \cline{2-5} 
 &
  t-SNE &
  Explainable &
  Local &
  \cite{dobrakowski2021interpretable, sushil2018patient,   minot2022interpretable, bhatt2021dice} \\ \hline
\multirow{2}{*}{Input-based} &
  Feature importance &
  Interpretable &
  Local &
  \cite{lindsay2021language, garcia2021biomedical,  ong2020machine, holderness2019analysis,   mellado2019assessing,     sushil2018patient, xue2019explainable, gin2022exploring} \\ \cline{2-5} 
 &
  Knowledge base/graph &
  Interpretable &
  Local &
  \cite{frisoni2021phenomena, mandalios2021enriching,   zhang2022unified, gu2020learning, teng2020explainable} \\ \hline
\multirow{3}{*}{Output-based} &
  Attention &
  Interpretable &
  Local &
  \cite{dong2021explainable, teng2020explainable,   duarte2018deep, zhang2018patient2vec,   trigueros2022explainable,     chen2022training, zhu2022using, ahmed2022eandc, ahmed2022multi,   luo2021applying, balagopalan2021comparing} \\ \cline{2-5} 
 &
  Evidence-based &
  Explainable &
  Local &
  \cite{de2022explainable} \\ \cline{2-5} 
 &
  Sentiment intensity score &
  Explainable &
  Local &
  \cite{boukobza2022deep} \\ \hline
\end{tabular}%
}
\caption{Explainable and interpretable artificial intelligence (XIAI) methods, their category, and scope}
\label{tab:1}
\end{table}

\subsubsection{Model-based XIAI}
Model-based XIAI focuses on describing how DL functions through the use of internal or external modules. The following section presents the most promising XAI and IAI methods that we have identified.

\textbf{LIME}. \citet{ribeiro2016model} introduced the local interpretable model-agnostic explanation (LIME) method. To produce explanations, LIME perturbs an instance, generates neighbouring data, and learns linear models within that neighbourhood. For instance \citet{zhang2019explainable} analyzed physician ordering for magnetic resonance imaging (MRI) scans and corresponding radiology reports (outcomes), by developing an RNN model integrated with a LIME module. The authors showed that it is possible to justify whether a patient should undergo MRI or not and if any brain abnormalities are likely to be detected, based on certain keywords from the physician order text entries. 

\textbf{SHAP}. Shapley Additive exPlanations (SHAP) is an algorithm used to obtain local, post-hoc explanations \cite{lundberg2017unified}. The SHAP method is a model agnostic approach, that allows to elucidate how individual input features contribute to model decisions. The method is based on deriving Shapley values from cooperative game theory \cite{shapley1997value}, which aims to quantify the contribution of each input, both individually as well as collectively in combination with all other inputs. \citet{naseem2022hybrid}introduced a Transformer-based encoder with a SHAP module for suicide risk identification. The SHAP module collected the contribution of important words that were most relevant to mental health and suicide.

\textbf{t-SNE}. t-SNE can effectively capture local interactions within data, although it is less effective in identifying long-range interactions \cite{van2008visualizing}. Using t-SNE, it is possible to visualize local clusters consisting of specific word embeddings, towards enhancing model explainability \cite{dobrakowski2021interpretable, sushil2018patient, minot2022interpretable, bhatt2021dice}.

Although less explored, model-based techniques may be promising for IAI (Table \ref{tab:1}). \citet{lu2021revealing} employed an opinion aggregator in combination with BERT, which involved k-means clustering to disentangle responses for COVID-19 questions. Another study developed a Bayesian network-derived causal graph to identify causal relationships between NLP variables extracted from mammography reports \cite{chen2021causal}. The authors fed the causal graph into a TabNet model and showed that they could improve model interpretability. Causal modelling can be a promising technique to perform global IAI. 

\subsubsection{Input-based XIAI}
Input-based XIAI methods focus on understanding how specific inputs influence network decisions. All input-based methods identified were either under the feature importance or knowledge base/graphs categories.

\textbf{Feature Importance}. \citet{lindsay2021language} extracted semantic, syntactic and paralinguistic features using NLP from spontaneous picture descriptions to identify cognitive impairments for Alzheimer’s disease (AD) detection. The authors quantified statistically significant (``generalizable'' ) NLP features, across two different languages. Subsequently, they performed binary (AD vs healthy) patient classification using DL and showed more accurate results when the generalizable features were used, against when all features were considered. 
\citet{holderness2019analysis} integrated clinically relevant multiword expressions during preprocessing which improved the accuracy and interpretability of their DL models (multilayer perceptrons and radial basis function DL) in identifying psychosis patients at risk of hospital readmission. \citet{mellado2019assessing} engineered clinically interpretable NLP-based features from topic extraction and clinical sentiment analysis to predict early readmission risk in psychiatry patients. In a different setting, \citet{xue2019explainable} designed a knowledge extraction framework from heterogeneous medical data sources, capable of generating an aggregated dataset to characterize diseases. Based on this dataset, they introduced an end-to-end DL-based medical multi-disease diagnosis system.  

\textbf{Knowledge graphs.}
Knowledge graphs (KGs) are employed to structure knowledge representations from either internal or external sources \cite{hamilton2020graph, wang2018describing, fu2020survey}. \citet{teng2020explainable} introduced an end-to-end KG-based framework for diagnosis code predictions. \citet{gu2020learning} proposed a novel method to solve complex question-answering from a medical dataset by combining KG information. The authors used Wikipedia as an external text source to extract documents related to questions and then mined triples (subject, predicate, object) from the documents to construct a KG. They subsequently extracted evidence graphs from the KG and embedded them into an Attention-based Graph Neural Network to perform complex question answering.

\subsubsection{Output-based XIAI}
Output-based XIAI methods focus on explaining/interpreting DL models by unravelling how internal computations within DL converge to output decisions.

\textbf{Attention Mechanisms}. The attention mechanism was initially designed to learn long-range interactions in NLP and improve machine translation \cite{bahdanau2014neural}. The attention mechanism allows to search for a set of positions in a source sentence and encourages the model to predict a target word based on the context vectors associated with these source positions and all the previously generated target words \cite{bahdanau2014neural}. Following attention, the self-attention mechanism was designed to further enhance the modelling of long-range interactions \citep{cheng2016long}. Within a self-attention layer, each position (known as query, key, and value) can attend to all positions in the output of the previous layer. Attention and self-attention mechanisms are the building blocks of Transformers. Transformer models have been recently applied in biomedical NLP \citep{chen2022training, zhu2022using} and image analysis \citep{papanastasiou2023attention}. Next to performance gains, attention mechanisms and Transformers have demonstrated their applicability in enhancing interpretability \cite{mascharka2018transparency}. Attention mechanisms were the most widely used XIAI techniques in our review.

\citet{dong2021explainable} designed a Hierarchical Label-wise Attention Network to perform medical coding. The proposed model was able to provide interpretations in the form of attention weights quantified from both words and sentences. \citet{duarte2018deep}  engineered attention mechanisms inside a neural network to derive death certifications through the analysis of diagnosis codes. Their method produced accurate and interpretable results and can potentially be transferrable to clinical settings. \citet{zhang2018patient2vec} incorporated a series of self-attention modules into an RNN architecture. They showed accurate and interpretable predictions of patients at risk for all-cause hospitalizations through the analysis of EHR data. \citet{chen2022training} trained BERT-based models via federated learning for ICD-10 coding. To interpret the model outputs, the authors added a label attention mechanism to the BERT model, which was able to display federated learning outcomes. 

\begin{figure}[]
\centering
\includegraphics[width=1\textwidth]{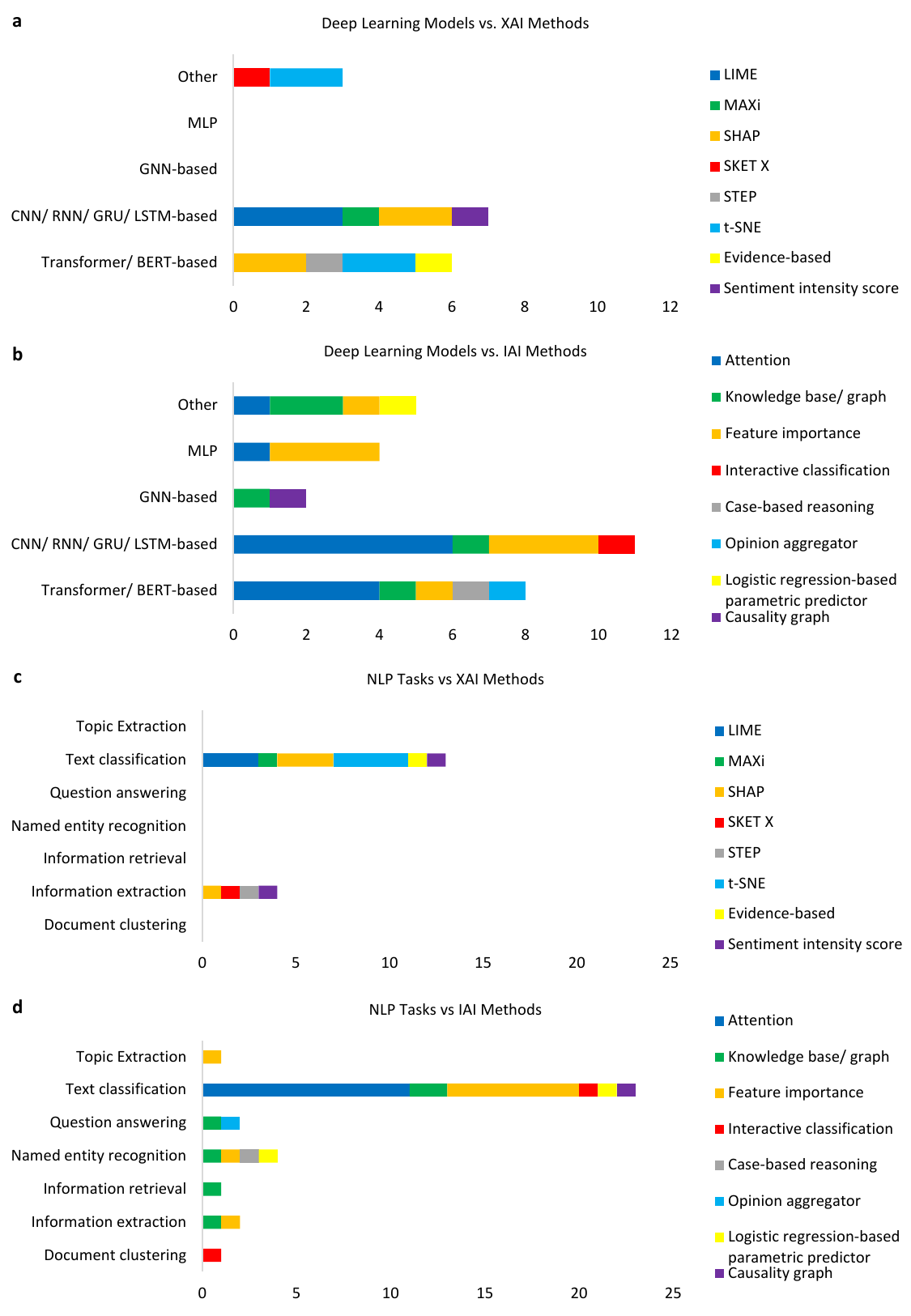}
\caption{\label{fig:3}Publication record showing combinations between (a) deep learning model and XAI method, (b) deep learning model and IAI method, (c) NLP task and XAI method and (d) NLP task and IAI method}
\end{figure}

\subsection{Deep learning}
Table \ref{tab:2} presents all DL models identified across papers. DL models were classified as 1) CNN/RNN/GRU/LSTM-based models; 2) Transformer/BERT-based models, featuring pre-trained language models; 3) Multilayer Perceptrons (MLPs); 4) Graph neural networks (GNNs) and; 5) other which includes models used once like unsupervised learning \cite{ferte2021automatic} and stacked denoising autoencoder \cite{sushil2018patient}.

The publication trends across DL models were examined (Figure \ref{fig:1}(c)). Figures \ref{fig:3}(a) and (b) illustrate DL model combinations with XAI and IAI, respectively. CNN/RNN/GRU/LSTM-based models were the most frequently used ($n = 17$), followed by Transformer/ BERT-based ($n = 14$), MLP ($n = 4$), GNN-based ($n = 2$) and other models ($n = 7$). Considering the period 2021-2022, Transformer/BERT-based models were the most dominant backbone architecture used. Although Transformer-based techniques were introduced in 2017 \cite{vaswani2017attention}, our analysis shows that their adoption in the healthcare domain was not immediately adopted.

\subsection{XIAI and NLP}
As shown in Table \ref{tab:2}, seven NLP tasks were identified: 1) text classification (which includes sentiment analysis \cite{mellado2019assessing}), 2) information extraction, including knowledge extraction \cite{marchesin2022empowering, xue2019explainable} and relation extraction \cite{zhang2022unified}, 3) named entity recognition, 4) question answering, 5) document clustering, 6) information retrieval and 7) topic extraction. Text classification was the main NLP task ($n = 35$), followed by information extraction ($n = 5$), named entity recognition ($n = 4$) and other 4 tasks ($n = 5$). 

Figure \ref{fig:3}(c-d) presents NLP tasks against XAI and IAI, respectively. Of note, IAI methods have been employed across all NLP tasks defined. In contrast, XAI methods have been exclusively used for text classification and information extraction.

\subsection{XIAI and medical tasks}
Medical tasks across all studies were extracted and categorized into sixteen categories (see Table \ref{tab:2}). Figure \ref{fig:4} illustrates the primary medical tasks, including medical diagnosis ($n = 10$), medical coding ($n =6$), medical decision making ($n=6$), mental health diagnosis ($n = 5$), disease classification ($n = 3$) and other ($n = 12$). 

Figure \ref{fig:4} (a-b) illustrates all medical tasks addressed using XAI and IAI methods, respectively. IAI methods were broadly applied to a more diversified set of medical tasks, against XAI methods which were mainly utilized in mental health diagnosis, medical diagnosis and decision making. 

\subsection{Evaluation metrics and open code}
In our review, only 3 out of all 42 articles involved dedicated evaluation processes and metrics of XIAI methods. These studies either involved qualitative evaluations by developers \cite{dong2021explainable} and end-users (i.e., clinicians) \cite{teng2020explainable}, or developed a quantitative explainability index \cite{farruque2021explainable}.

Limited code availability (only 5 out of 42 studies provided code access) was identified across publications, which restricts the ability to assess if these XIAI methods can be replicated on public or future benchmark datasets \cite{ferte2021automatic,dong2021explainable,garcia2021biomedical,frisoni2021phenomena,bhatt2021dice}. 

\subsection{Datasets}
Finally, methods were categorized based on whether they were developed using \textit{public datasets} (open source and accessible), or \textit{private datasets} (otherwise). According to Table \ref{tab:2}, the majority of papers(n = 32 out of 42 papers) used private data and only 10 papers analyzed public datasets. EHR from hospitals was the most prevalent type of private data analyzed (Table \ref{tab:2}). The most widely used public dataset was MIMIC ($n = 5$), which is a large clinical database consisting of multi-modal (text and imaging) data (Table \ref{tab:2}).

\begin{figure}[htb]
\centering
\includegraphics[width=0.85\textwidth]{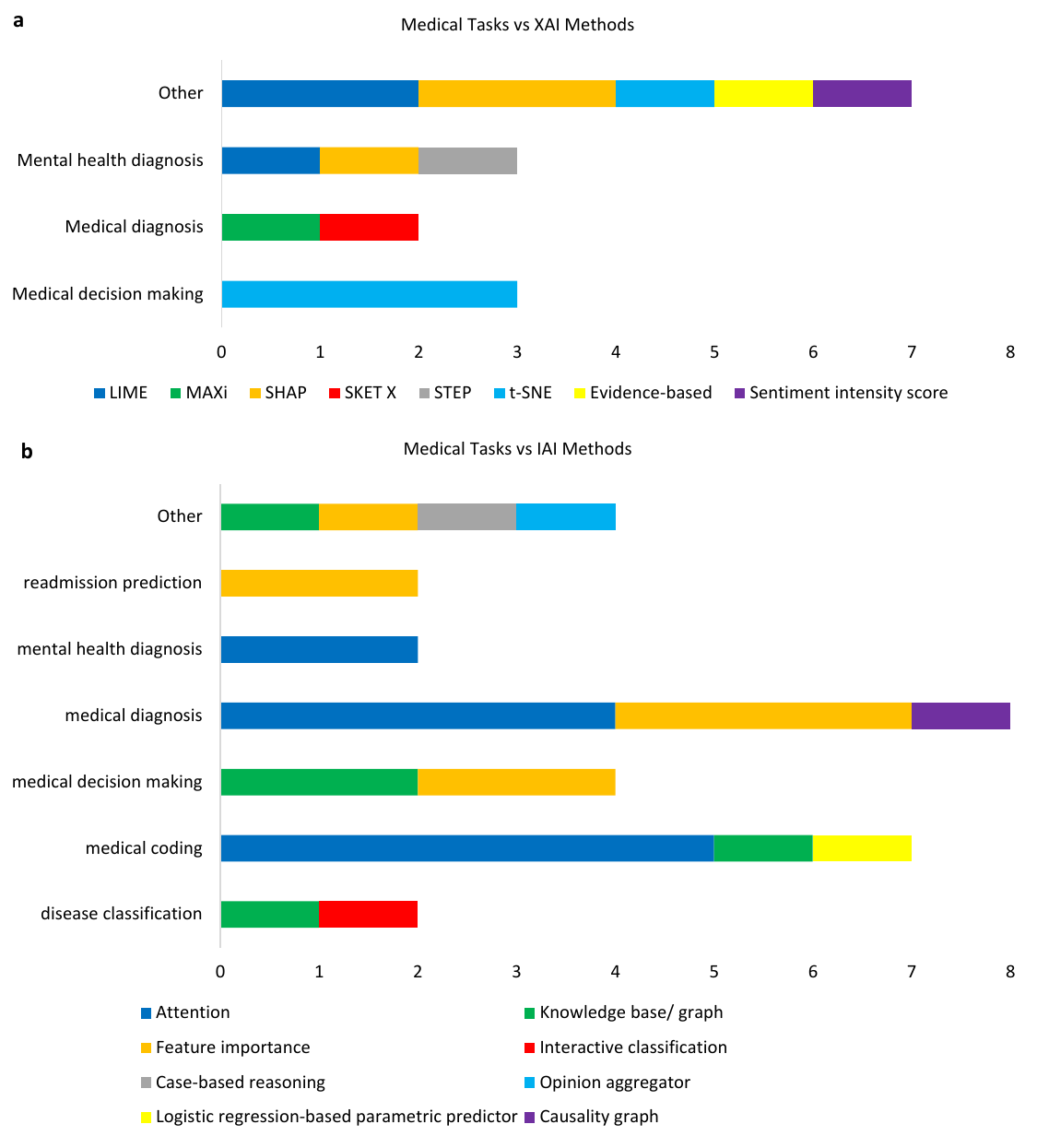}
\caption{\label{fig:4}Publication record showing combinations between medical task and (a) XAI and (b) IAI method.}
\end{figure}

{
\scriptsize
\begin{landscape}
\begin{longtable}[c]{|c|p{4cm}|p{6cm}|l|l|c|}
\hline
\textbf{DL   Model} &
  \multicolumn{1}{c|}{\textbf{Medical task category}} &
  \multicolumn{1}{c|}{\textbf{Medical task}} &
  \multicolumn{1}{c|}{\textbf{NLP task}} &
  \multicolumn{1}{c|}{\textbf{Dataset}} &
  \multicolumn{1}{c|}{\textbf{Article}} \\ \hline
\endfirsthead
\multicolumn{6}{c}%
{{\bfseries Table \thetable\ continued from previous page}} \\
\hline
\textbf{DL   Model} &
  \multicolumn{1}{c|}{\textbf{Medical task category}} &
  \multicolumn{1}{c|}{\textbf{Medical task}} &
  \multicolumn{1}{c|}{\textbf{NLP task}} &
  \multicolumn{1}{c|}{\textbf{Dataset}} &
  \multicolumn{1}{c|}{\textbf{Article}} \\ \hline
\endhead
\multirow{14}{*}{\makecell[c]{Transformer/\\BERT-based}} &
  Medical entity disambiguation &
  Biomedical named entity   disambiguation &
  NER &
  ClinWikiNED &
  \cite{garcia2021biomedical} \\ \cline{2-6} 
 &
  Medical decision making &
  Chinese medicine instruction   parsing &
  \makecell[l]{NER; \\Information extraction} &
  Private dataset &
  \cite{zhang2022unified} \\ \cline{2-6} 
 &
  Medical bias mitigation &
  Reducing genderization &
  Text classification &
  MIMIC-III &
  \cite{minot2022interpretable} \\ \cline{2-6} 
 &
  Medical decision making &
  Drug indication classification &
  Text classification &
  \makecell[l]{curated drug \\indication corpus} &
  \cite{bhatt2021dice} \\ \cline{2-6} 
 &
  Medical research &
  Accelerate COVID-19 research &
  Question Answering &
  Private dataset &
  \cite{lu2021revealing} \\ \cline{2-6} 
 &
  Management of health services &
  Automatically scoring   request for proposals (RFP) &
  \makecell[l]{Information extraction; \\Text classification} &
  Private dataset &
  \cite{maji2021interpretable} \\ \cline{2-6} 
 &
  Medical note generation &
  Radiology report generation   assistance &
  NER &
  MIMIC-CXR &
  \cite{amador2021case} \\ \cline{2-6} 
 &
  Medical coding &
  Medical coding &
  Text classification &
  Private dataset &
  \cite{chen2022training} \\ \cline{2-6} 
 &
  Medical diagnosis &
  Long-term COVID eﬀects &
  Text classification &
  Private dataset &
  \cite{zhu2022using} \\ \cline{2-6} 
 &
  Medical fake news detection &
  Fake news detector to covid-19 &
  Text classification &
  Private dataset &
  \cite{de2022explainable} \\ \cline{2-6} 
 &
  Mental health diagnosis &
  Suicide risk identification &
  Text classification &
  Private dataset &
  \cite{naseem2022hybrid} \\ \cline{2-6} 
 &
  Medical diagnosis &
  Chronic cough &
  Text classification &
  Private dataset &
  \cite{luo2021applying} \\ \cline{2-6} 
 &
  Medical diagnosis &
  Alzheimer’s disease &
  Text classification &
  Private dataset &
  \cite{balagopalan2021comparing} \\ \cline{2-6} 
 &
  Mental health diagnosis &
  Depression symptoms detection   (DSD) &
  Text classification &
  Private dataset &
  \cite{farruque2021explainable} \\ \hline
\multirow{15}{*}{\makecell[c]{CNN/\\RNN/\\GRU/\\LSTM-based}} &
  Medical coding &
  Medical coding &
  Text classification &
  MIMIC-III &
  \cite{teng2020explainable} \\ \cline{2-6} 
 &
  Medical diagnosis &
  Ischemic stroke &
  Text classification &
  Private dataset &
  \cite{ong2020machine} \\ \cline{2-6} 
 &
  Disease classification &
  Predict adherence to ACR   guidelines &
  Text classification &
  Private dataset &
  \cite{zhang2019explainable} \\ \cline{2-6} 
 &
  Medical coding &
  Causes of death &
  Text classification &
  Private dataset &
  \cite{duarte2018deep} \\ \cline{2-6} 
 &
  Medical diagnosis &
  Diagnosis &
  Text classification &
  Private dataset &
  \cite{zhang2018patient2vec} \\ \cline{2-6} 
 &
  Medical decision making &
  Entrustment assessment &
  Text classification &
  Private dataset &
  \cite{gin2022exploring} \\ \cline{2-6} 
 &
  Medical diagnosis &
  Cancer pathology &
  Text classification &
  Private dataset &
  \cite{martina2020classification} \\ \cline{2-6} 
 &
  Survival detection &
  Discrete-time survival estimates &
  Text classification &
  Private dataset &
  \cite{thorsen2022discrete} \\ \cline{2-6} 
 &
  Medical coding &
  ICD multi-label classifcation &
  Text classification &
  \makecell[l]{MIMIC-III; \\Spanish Osa Corpus \\(Private dataset)} &
  \cite{trigueros2022explainable} \\ \cline{2-6} 
 &
  Management of health services &
  Automatically scoring   request for proposals (RFP) &
  \makecell[l]{Information extraction; \\Text classification} &
  Private dataset &
  \cite{maji2021interpretable} \\ \cline{2-6} 
 &
  Medical diagnosis &
  Disease diagnosis &
  \makecell[l]{Information extraction; \\Text classification} &
  Private dataset &
  \cite{xue2019explainable} \\ \cline{2-6} 
 &
  Monitor public sentiments during pandemics &
  COVID-19 pandemic &
  \makecell[l]{Information extraction; \\Text classification} &
  Private dataset &
  \cite{boukobza2022deep} \\ \cline{2-6} 
 &
  Telehealth services &
  Telehealth services &
  Text classification &
  Private dataset &
  \cite{ozyegen2022word} \\ \cline{2-6} 
 &
  Mental health diagnosis &
  Internet-delivered psychological   treatment (IDPT) &
  Text classification &
  Private dataset &
  \cite{ahmed2022eandc} \\ \cline{2-6} 
 &
  Mental health diagnosis &
  Depression &
  Text classification &
  Amazon Mechanical Turk &
  \cite{ahmed2022multi} \\ \hline
\multirow{2}{*}{} &
  Mental health diagnosis &
  Depression &
  Text classification &
  Private dataset &
  \cite{uddin2022depression} \\ \cline{2-6} 
 &
  Disease classification &
  Diabetes-related biomedical literature exploration &
  \makecell[l]{Document clustering; \\Text classification} &
  Private dataset &
  \cite{ahne2022improving} \\ \hline
\multirow{2}{*}{GNN-based} &
  Medical diagnosis &
  Breast cancer &
  Text classification &
  Private dataset &
  \cite{chen2021causal} \\ \cline{2-6} 
 &
  Medical   question-answering &
  Medical QA &
  Question Answering &
  Head-QA &
  \cite{gu2020learning} \\ \hline
\multirow{4}{*}{MLP} &
  Medical diagnosis &
  Language impairment in   Alzheimer’s disease &
  Text classification &
  Private dataset &
  \cite{lindsay2021language} \\ \cline{2-6} 
 &
  Readmission prediction &
  Psychiatric readmission prediction &
  Text classification &
  Private dataset &
  \cite{holderness2019analysis} \\ \cline{2-6} 
 &
  Readmission prediction &
  Psychiatric readmission risk   prediction &
  \makecell[l]{Text classification; \\Topic extraction} &
  Private dataset &
  \cite{mellado2019assessing} \\ \cline{2-6} 
 &
  Medical diagnosis &
  Alzheimer’s disease &
  Text classification &
  Private dataset &
  \cite{balagopalan2021comparing} \\ \hline
\multirow{7}{*}{Other} &
  Medical coding &
  Patient medical conditions &
  \makecell[l]{NER; \\Text classification} &
  Private dataset &
  \cite{ferte2021automatic} \\ \cline{2-6} 
 &
  Medical decision making &
  Clustering of patients’ visits &
  Text classification &
  Private dataset &
  \cite{dobrakowski2021interpretable} \\ \cline{2-6} 
 &
  Medical coding &
  Medical coding &
  Text classification &
  \makecell[l]{MIMIC-III-50; \\MIMIC-III-shielding; \\MIMIC-III} &
  \cite{dong2021explainable} \\ \cline{2-6} 
 &
  Medical decision making &
  Esophageal achalasia &
  Information retrieval &
  Private dataset &
  \cite{frisoni2021phenomena} \\ \cline{2-6} 
 &
  Disease classification &
  Cardiovascular diseases &
  Text classification &
  OHSUMED &
  \cite{mandalios2021enriching} \\ \cline{2-6} 
 &
  Medical decision making &
  Patient mortality prediction;   Primary diagnostic category prediction; Primary procedural category   prediction; Gender prediction &
  Text classification &
  Private dataset &
  \cite{sushil2018patient} \\ \cline{2-6} 
 &
  Medical diagnosis &
  Pathology &
  Information extraction &
  Private dataset &
  \cite{marchesin2022empowering} \\ \hline
\caption{All the articles grouped based on the XIAI method along with the associated medical task, NLP task and the dataset used for.}
\label{tab:2}\\
\end{longtable}
\end{landscape}
}

\section{Discussion}\label{sec:4.1}
\subsection{Challenges and Opportunities for clinical translation}
In this section we discuss the key opportunities and challenges surrounding the clinical translation of XIAI. XIAI presents significant opportunities such as leveraging attention mechanisms to combine and interpret
multi-modal information (text, images, genetic data, clinical history) for personalized medicine and combining DL with causal modelling towards further enhancing inherent interpretability. However, our findings also suggest that all these methods require strong in-house
technical expertise to infer XIAI. Other key challenges include the lack of established best practices for XIAI selection based on data and problem type, as
well as the unmet need for systematic evaluation methods and high-quality
benchmarks. In the following paragraphs, we expand on these challenges and opportunities, identified from the perspective of clinical translation.

Based on our literature review, there is no previous survey focusing on these topics. Considering their early development phase, more extensive studies are needed in the future to develop XIAI methods that will allow ``global'' (end-to-end) interpretations that go beyond visualization maps, devise more transparent XIAI that will require less technical oversight and design XIAI evaluation metrics that are critical to establish best practices for XIAI
selection. The following points can therefore guide future work and systematic reviews towards strengthening and democratizing XIAI in healthcare NLP.

\textbf{Challenges:}
The majority of publications focused on local XIAI, with only 5 papers involving a global XIAI approach \cite{chen2021causal, ferte2021automatic, lu2021revealing, amador2021case, ahne2022improving}. Moving beyond local XIAI (which provides relatively limited insights based on specific inputs or features) towards ``globally'' enhancing the transparency of the entire modelling process might be necessary to reduce the requirement of in-house technical expertise and to develop readily translational XIAI methods for end-users. 

In our review, the evaluation of XIAI methods was substantially limited (only 3 papers involved XIAI evaluation metrics). Developing robust evaluation metrics for XIAI is challenging given the limited availability of ground truths in healthcare data \cite{papanastasiou2023attention} and the complexity of rapidly evolving models (like LLMs) \cite{ouyang2022training}. The majority of papers analyzed private data (32 papers) and did not provide source code (37 papers). Legal implications around private data pose another critical challenge in terms of democratizing data and assessing reproducibility for these XIAI techniques. Hence, it is particularly challenging to assess the reproducibility of these XIAI NLP methods. Significant variability in the literature was observed across model-, input- and output-based methodologies used (Table \ref{tab:1}) and NLP and medical tasks addressed (Figures \ref{fig:3} and \ref{fig:4}). Each of these method types presents different advantages and drawbacks (see subsection \ref{sec:4.2}). It is therefore substantially challenging to develop generalization frameworks regarding which XIAI method is most optimal across different NLP and medical applications. In that context, there is an unmet yet pivotal need to develop benchmark data/metrics to extensively evaluate XIAI methods across different NLP and medical tasks, with associated open-access codes.

While attention-based IAI holds promise, it primarily relies on visualizing important attention weights as heat maps \cite{bahdanau2014neural, cheng2016long}. Heat maps may reveal patterns in the way the model prioritizes specific types of words or grammatical structures, providing some insight into which parts of the processed text are most influential. However, heat maps highlight what the model "looks at," but not necessarily how it interprets that information. Moreover, attention weights do not linearly correlate with model outcomes \cite{sun2021interpreting, balagopalan2021comparing}.  These two main limitations currently compromise attention-based IAI.
The adoption of current XIAI in healthcare and industry systems is therefore challenging, given the limited access to global XIAI techniques and the absence of robust XIAI metrics, ground truths and benchmark data/ studies, which form an important barrier to their systematization.

XIAI systematization and deployment necessitate further thorough work. A possible practical mitigation to accelerate the systematization of robust and transparent XIAI developments for NLP in healthcare is to bring ``humans into the DL loop'' \cite{mosqueira2023human, wu2022survey}: domain experts, end-users, policymakers and patients will be able to contribute to the XIAI method design, development and evaluation. This collective approach can potentially lead to the emergence of robust and ready-to-use XIAI methods across different NLP and medical tasks.

\textbf{Opportunities:}
Attention mechanisms were the most diversified XIAI techniques in terms of DL used (Figure \ref{fig:3} a, b) as well as NLP (Figure \ref{fig:3} c, d) and medical tasks (Figure \ref{fig:4}) addressed. It is noteworthy that although Transformers were the dominant DL backbones in the period 2021-2022, individual attention mechanisms were the most frequently used IAI technique. Attention and self-attention mechanisms have also recently been used in combination with CNN models for medical image analysis, as lighter alternatives of Transformers \cite{papanastasiou2023attention}. This demonstrates their diverse applicability across data domains and tasks. A major opportunity identified is the versatility of attention mechanisms to be combined with multiple DL structures (e.g., CNN \cite{trigueros2022explainable}, RNN \cite{zhang2018patient2vec}, BERT \cite{chen2022training} and full Transformers) \cite{balagopalan2021comparing}. Another translational opportunity identified via the use of attention mechanisms is to simultaneously model and interpret information from variable multi-modal data (e.g., text, images, genetics, clinical history) \cite{papanastasiou2023attention}. Based on the previous, developing multi-modal XIAI can potentially support personalized medicine which benefits from combining patient-level information from multiple sources \cite{salvi2023multi}. Despite the challenges associated with attention-based heat maps described above, combining information from multiple sources e.g., images and text, can potentially enhance interpretability. A characteristic example are DL models developed for MIMIC data analysis, in which features extracted from X-ray images and radiology reports are combined to create a unified representation that fuses information from both modalities \cite {teng2020explainable, dong2021explainable, amador2021case, dobrakowski2021interpretable, minot2022interpretable}. Attention heat maps can highlight which parts of the image and text were most influential in the model decision. Combining image and text data can offer insights into a model's decision-making that go beyond the limitations of single-modal (text-only) attention heat maps.  

A clear trend towards IAI against XAI was observed across studies, particularly in the period 2021-2022. While our review highlights the benefits of using IAI against XAI, their combination can be useful as an auxiliary assessment to cross-evaluate IAI and XAI outcomes. For instance, the SHAP method that can provide both global and local explanations \cite{lundberg2017unified}, has been effectively combined with a Transformer encoder for suicide risk prediction \cite{naseem2022hybrid}. Although this work focused on the SHAP method to perform XAI, future work could aim to also co-extract IAI information from specific attention mechanisms within Transformers.

Causality is an emerging topic in DL which aims to improve model interpretability, fairness and generalization \cite{scholkopf2021toward,pearl2010introduction}. The fundamental aim of causal DL is to unravel causal relationships between variables which determine the model's decision-making process \cite{scholkopf2021toward}. Understanding how the data are causally related can help us design better DL models. This leads to more reliable predictions and a deeper understanding of how our models work \cite{feder2022causal}. Next to identifying causal paths between data, central to causality is also to understand the relationships between cause and effect \cite{pearl2010introduction}. Understanding cause and effect is crucial for many important decisions. In clinical trials for example, doctors need to know if a new drug actually improves patient outcomes (not just whether there is a statistical correlation). In our review, a Bayesian Network-derived causal graph has been fed into a TabNet, showing promising IAI results \cite{chen2021causal}. Further work in this field can significantly enhance IAI in NLP for healthcare. We endeavour to inspire and guide relevant benchmark studies to thoroughly examine XIAI in terms of strengthening NLP applications in healthcare. For example, leveraging causal mechanisms to better understand how foundation models such as Transformers interact with data or prompts, may be a critical path forward. 

\subsection{The three XIAI paradigms}\label{sec:4.2}

\textbf{Model-based XIAI methods}, such as SHAP \cite{lundberg2017unified}, LIME \cite{ribeiro2016model}, and t-SNE \cite{van2008visualizing}, offer important advantages by providing both global and local DL interpretability options. These methods are designed to be transparent and accessible as ready-to-use entities thus, potentially being able to democratize XIAI tools for a wide audience \citep{thorsen2022discrete}. Intuitive visualizations can to some extent elucidate complex model predictions, enhancing trust and facilitating the communication of findings. However, they also have drawbacks, such as the instability of methods like LIME, which is prone to minor data variations that can affect XAI reliability \citep{rudin2019stop, balagopalan2021comparing}. Moreover, relying on separate modules for XIAI may be subject to misinterpretations, in case these modules have inherent biases or limitations. Hence, despite their ease of use, they require technical expertise to evaluate any biases or implementation incompatibilities.

\textbf{Input-based XIAI methods} allow for DL model interpretation by leveraging specific input feature importance \cite{lindsay2021language, garcia2021biomedical, ong2020machine} and medical KG \cite{frisoni2021phenomena, mandalios2021enriching, zhang2022unified}, to discern how different inputs influence model decisions. These methods are intuitive and accessible to computer scientists and NLP researchers, due to their relative ease of being combined with DL architectures. In that context, the incorporation of ``important features'' or medical KGs into AI can enrich models with domain-specific insights, which in turn allows them to interpret even complex medical concepts that are common in clinical practice \citep{frisoni2021phenomena}.

Input methods based on feature importance requires the incorporation of techniques that are not readily transparent to non-technical medical professionals (end-users). Hence, it may not always be straightforward for end-users to understand ``how an important feature was derived and/or evaluated'' \citep{lindsay2021language, holderness2019analysis, mellado2019assessing}. The effectiveness of KG methods relies on medical professionals with expertise and capacity to develop comprehensive medical KG, posing a challenge if such expertise is scarce. Integrating medical KG into DL presents challenges in terms of ontology construction as well as knowledge extraction  \citep{teng2020explainable}, which are labour-intensive techniques.

\textbf{Output-based XIAI methods}, are important to interpret DL outputs and can offer computational insights through mechanisms like attention \cite{mascharka2018transparency}. Output-based methods focus on explaining/interpreting DL models by uncovering how internal computations within DL converge to output decisions, which can be mainly useful to computer scientists/ modellers. Nevertheless, attention mechanism-based IAI faces debates \cite{serrano2019attention, pruthi2020learning, wiegreffe2019attention, vashishth2019attention, brunner2019identifiability}, as high attention weights don't necessarily linearly correlate with model predictions \cite{sun2021interpreting}. This can lead to ambiguity while emphasizes the need for further research on IAI methods and their evaluation.

\subsection{Inspired by the future}
\textbf{Go Large:} Recently, LLMs have attracted significant attention in AI \cite{openai, brown2020language, achiam2023gpt}. The intersection of LLMs and healthcare creates unique opportunities towards designing future studies, from drug discovery to personalized diagnosis and treatment \cite{mesko2023imperative, tu2024towards, singhal2023publisher}. Healthcare data analysis is one of the high gain-high risk domains for LLMs \cite{ouyang2022training}. One of the limitations of LLMs is that they sometimes tend to ``hallucinate'' results \cite{openai, brown2020language, achiam2023gpt, openai, brown2020language, achiam2023gpt}. This can add considerable barriers to their utilization in healthcare settings  \cite{mesko2023imperative, tu2024towards, singhal2023publisher}. As discussed in subsection \ref{sec:4.1}, causal inference can be an effective solution towards enhancing IAI \cite{scholkopf2021toward}. Although designing causal logic inside LLMs is challenging due to their architectural complexity and the fact that models are already ``trained'' on a causal-agnostic mode, there have been recent attempts which aim to develop causal reasoning between prompts and responses \cite{jin2023cladder,hobbhahn2022investigating}.

\textbf{Go Small:} An ongoing discussion in the community is whether LLMs or domain-specific smaller models can be more robust solutions for healthcare data \cite{wang2023large}. Recently, smaller-parameter domain-specific LMs have outperformed larger LMs when examined on clinical notes from large public databases (MIMIC) \cite{hernandez2023we}. This approach has several possible benefits: a) small-parameter LMs can be trained using in-house computational capabilities, which b) minimizes the risks associated with transmitting sensitive patient data to cloud-based or external servers (thus, adhering to privacy regulations). Moreover, c) causal DL techniques can be more optimally designed and validated since they can be trained on domain-specific data from scratch.

\section{Conclusions}

There is a growing trend of using IAI techniques alongside XAI, with attention mechanisms emerging as the most dominant approach for interpretation. Our analysis reveals a major opportunity: leveraging attention mechanisms to combine and interpret multi-modal data (text, images, genetics, etc.), potentially leading to advances in XIAI approaches for personalized medicine solutions \cite{papanastasiou2023attention}. Another promising XIAI opportunity lies in combining DL with causality, which can further enhance inherent interpretability. While the benefits of model-, input-, and output-based XIAI have been demonstrated, our findings suggest that the drawbacks of all these methods is that they require in-house technical expertise to infer XIAI. The key challenges include the lack of established best practices for XIAI selection based on data and problem type, as well as the unmet need for systematic evaluation metrics and high-quality benchmarks. Moreover, to effectively adopt DL in healthcare NLP, moving from ``local'' to ``global'' XIAI techniques that can inform the entire modelling process is crucial. Finally, our discussion encourages the integration of XIAI into the ongoing development of LLMs and domain-specific smaller language models.

Our results indicate that XIAI adoption in healthcare is not possible without dedicated in-house technical expertise. Bringing ``humans into the DL loop'' (domain experts, end-users, policymakers and patients) may be an effective solution to collectively design, develop and evaluate ready-to-use XIAI methods across different NLP and medical tasks. Despite the challenges identified, the XIAI techniques detailed in this review offer a valuable foundation for further research and benchmark studies. This will ultimately lead to enhanced inherent interpretability and facilitate the use of complex NLP algorithms in healthcare settings.

\bibliographystyle{elsarticle-num-names} 
\bibliography{references}
\end{document}